\documentclass[10pt,twocolumn]{article} 
\usepackage{simpleConference}
\usepackage{times}
\usepackage{graphicx}
\usepackage{amssymb}
\usepackage{url,hyperref}
\usepackage{amsmath}
\usepackage{multirow}
\usepackage{diagbox}
\usepackage[noend]{algpseudocode}
\usepackage{algorithmicx,algorithm}

\begin{document}

\title{Bi-Skip: A Motion Deblurring Network Using Self-paced Learning}

\author{Yiwei Zhang$^1$, Chunbiao Zhu$^1$, Ge Li$^1$, Yuan Zhao$^2$, Haifeng Shen$^2$\\
Peking University$^1$, AI Labs,Didi Chuxing$^2$\\
{yuriyzhang@pku.edu.cn}, {gli@pkusz.edu.cn}, {zhuchunbiao@pku.edu.cn}\\
{zhaoyuanjason@didiglobal.com}, {shenhaifeng@didiglobal.com}\\
\today
}

\maketitle
\thispagestyle{empty}

\begin{abstract}
A fast and effective motion deblurring method has great application values in real life. This work presents an innovative approach in which a self-paced learning is combined with GAN to deblur image. First, We explain that a proper generator can be used as deep priors and point out that the solution for pixel-based loss is not same with the one for perception-based loss. By using these ideas as starting points, a Bi-Skip network is proposed to improve the generating ability and a bi-level loss is adopted to solve the problem that common conditions are non-identical. 
Second, considering that the complex motion blur will perturb the network in the training process, a self-paced mechanism is adopted to enhance the robustness of the network.
Through extensive evaluations on both qualitative and quantitative criteria, it is demonstrated that our approach has a competitive advantage over state-of-the-art methods.
\end{abstract}

\section{Introduction}
\begin{figure}[t]
\centering
\includegraphics[width=8.4cm]{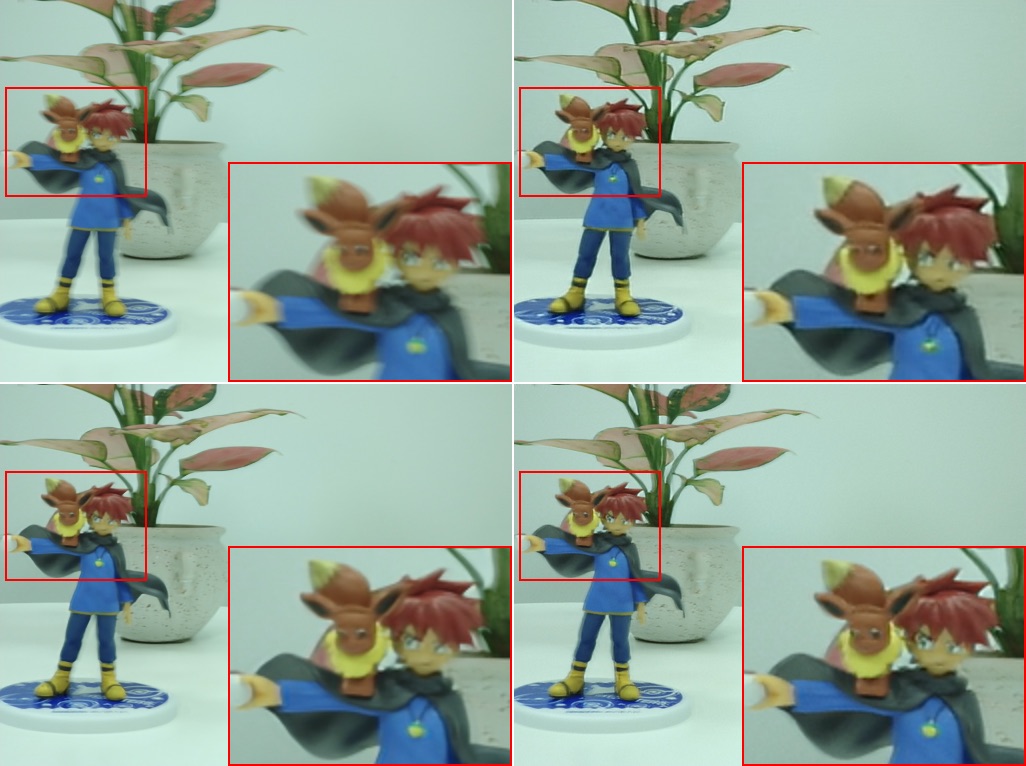}
\caption{Upper Left: blurred image, upper right: by \cite{Kupyn2017DeblurGAN}, bottom left: by \cite{tao2018scale}, bottom right: ours.}
\label{fig.first-show}
\end{figure}

Motion blur is a phenomenon caused by the relative motion of the camera and the target during the exposure time. Removing motion blur (see figure \ref{fig.first-show}) from single photograph has been an intractable problem though there are many approved works. However, it is an ill-posed problem to restore the sharp image when only the blurred image is given. To solve this blind equalization, conventional methods\cite{fergus2006removing,tai2011richardson,shan2008high,krishnan2009fast,cho2009fast,freeman2009understanding,xu2010two,pan2014motion,pan2016blind,zhang2015image} are based on the idea that regularizing the potential distributions of sharp image. In general, these methods deblur images through two steps including kernel estimation and non-blind deconvolution. On the one hand, the kernel estimation\cite{xu2010two,whyte2012non,pan2016blind,pan2014motion,pan2016robust,pan2016robust,mai2015kernel} is always a hard work because of the influence of variational motion and depth information which increase the difficulty of solving the inverse problem. On the other hand, the non-blind deconvolution\cite{richardson1972bayesian,wiener1949extrapolation,cho2011handling,zhang2017learning} often suffers ringing artifacts no matter how precisely the blur kernel is estimated.

Recently, the superior performance based upon deep learning attracts attention from all fields and some works\cite{Xu2014Deep,ronneberger2015u,Sun2015Learning,Wieschollek2017Learning,nah2017deep,Kupyn2017DeblurGAN,ulyanov2017deep,harmeling2010space} have applied it to image restoration. At the beginning, the basic idea to deblur\cite{Xu2014Deep} is not different from the family of those conventional methods but through using Convolutional Neural Network (CNN) in the kernel estimation and non-blind deconvolution. However, it seems that the fatal drawbacks of deblurring model mentioned above are not overcome. Namely, a more sensible model should have ability to reduce the intermediate steps which often introduce extraordinary artifacts. Thinking further, applying an encoder-decoder to restore the image seems to be a proper way to reduce the complexity of process.

By understanding U-Net\cite{ronneberger2015u} and Skip-Net\cite{Mao2016Image,ulyanov2017deep}, we know that networks can learn more details from the input when shallow features and deep features are concatenated together. Inspired by this idea, a Bi-Skip network is proposed to improve the generating ability. Besides, considering that the complex motion blur will perturb the network in the training process, a self-paced mechanism is introduced to enhance the robustness of the network.

Our contributions are given in three aspects. First, we propose a Bi-Skip network, which obtains competitive results with faster speed in image deblurring. Second, we point out that the solution for pixel-based loss is not same with the one for perception-based loss and a bi-level loss is adopted to solve this non-identical problem. Thirdly, a self-paced learning mechanism is introduced in our training process. And according to our knowledge, this is the first time to use self-paced mechanism to optimize the generative adversarial network.

\section{Related Work}
\subsection{Motion Deblurring}
Conventional motion deblurring methods are divided into two classes that are uniform and non-uniform. Derived from the convolution model, almost all previous works perform the non-blind deconvolution based on the estimated kernel. Early works\cite{fergus2006removing,shan2008high,cho2009fast,freeman2009understanding,xu2010two} mainly focus on uniform deblurring. As explained, these methods introduce the priors to regularize the distributions of sharp image. Nevertheless, neither heavy-tail\cite{fergus2006removing} nor hyper-laplacian\cite{krishnan2009fast} can generalize the distribution of real images. To improve the robustness of deblurring, Cho\cite{cho2011handling} and Xu\cite{xu2010two} introduce the edge information to regularize the optimal equation. Furthermore, Xu\cite{xu2010two,xu2013unnatural} proposes a $L_0$-norm regularizer which can extract more salient structures. From the point of probability, Levin\cite{freeman2009understanding} comprehensively elaborates the problem in the optimization and proposes an alternative $MAP_{k}$. Due to the limitation of uniform hypothesis, the optimal models are often in the cases of under fitting. In order to overcome this drawback, Harmeling\cite{harmeling2010space} and Whyte\cite{whyte2012non} contribute the solution in non-uniform deblurring. While, there exit two fatal drawbacks: the pool precision of blur kernels and the ringing artifacts of the non-blind deconvolution. By utilizing Deep Neural Network (DNN), impressive works\cite{Wieschollek2017Learning,nah2017deep,Kupyn2017DeblurGAN,Sun2015Learning} have been done in the image deblurring. Recently, Nah\cite{nah2017deep} proposes a kernel-free end-to-end approach to deblur images and Kupyn\cite{Kupyn2017DeblurGAN} suggests a blur-to-sharp translation by using cGAN. Also, Tao\cite{tao2018scale} proposes a scale-recurrent network for image deblurring.
\subsection{Self-paced Learning}
The philosophy under self-paced learning (SPL) is to simulate the learning principle of humans/animals, which generally starts by learning easier aspects of a learning task, and then gradually takes more complex examples into training \cite{khan2011humans}. Formally, the SPL model can be expressed as:
\begin{equation}
\operatorname*{min}\limits_{\mathbf{w},\mathbf{v}\in[0,1]^n}\mathbf{E}(\mathbf{w},\mathbf{v},\lambda) = \sum_{i=1}^{n}v_i\mathcal{L}(y_i,g(x_i,\mathbf{w}))+f(\mathbf{v},\lambda)
\end{equation} Given a training dataset $\mathcal{S}=\{(x_i,y_i)\}_{i=1}^{n}$, in which $x_i$ and $y_i$ denote the $i^{th}$ observed sample and its label respectively. $\mathcal{L}$ is a loss function which calculates the cost between the estimated label $g(x_i,\mathbf{w})$ and its ground truth. $\mathbf{w}$ represents the model parameter inside the decision function $g$. $\mathbf{v}=[v_1,\ldots,v_n]$ is latent weight variable for training samples. The ‘age’ $\lambda$ controls the learning space. When $\lambda$ is small, only 'easy' samples are considered into training. As $\lambda$ grows, more samples with larger losses will be gradually appended to train a more mature model. And $f(\mathbf{v},\lambda)$ is the self-paced regularizer which follows the definition: 1) $f(v,\lambda)$ is convex with respect to $v\in[0,1]$; 2) $v^*(l,\lambda)$ is monotonically decreasing with respect to $l$, and it holds that $lim_{l\rightarrow0}v^*(l,\lambda)=1$, $lim_{l\rightarrow\infty}v^*(l,\lambda)=0$; 3) $v^*(l,\lambda)$ is monotonically increasing with respect to $\lambda$, and it holds that $lim_{\lambda\rightarrow\infty}v^*(l,\lambda)\le1$, $lim_{\lambda\rightarrow0}v^*(l,\lambda)=0$. Furthermore, Meng \cite{meng2015objective} proves that the solving strategy on SPL accords with a majorization minimization algorithm implemented on a latent objective function. And Li \cite{li2017self} propose a self-paced convolutional neural networks to improves CNNs.

\subsection{Encoder-Decoder Networks}
The idea of Autoencoder is that high-dimensional data can be converted to low-dimensional codes by training a multilayer network to reconstruct high-dimensional input vectors.
That is, the features learned from networks should own the ability to describe the inputs as much as possible. Based on this idea, Denosing Autoencoder (DAE)\cite{vincent2010stacked} and Variational Autoencoder (VAE)\cite{kingma2013auto} have been used in the area of image restoration. 

At the beginning, U-Net is proposed for image segmentation. It consists of a contracting path and an expansive path. The contracting path follows down-sampling steps and the expansive path follows up-sampling steps. Due to the symmetry of U-Net, it can localize the pixels, which makes features concatenating easily. The Sip-Net proposed by Mao\cite{Mao2016Image} mainly focuses on image denoising and super-resolution. Deriving from U-Net, Skip-Net links convolutional and deconvolutional layers by skip-layer connections. Furthermore, Ulyanov\cite{ulyanov2017deep} points that, untrained deep convolutional generators can be used to replace the surrogate natural prior (TV norm) to gain dramatically improved results.  However, different from other image restoration, deblurring is still hard when directly using encoder-decoder. 

\begin{figure*}[t]
\centering
\includegraphics[width=17.8cm]{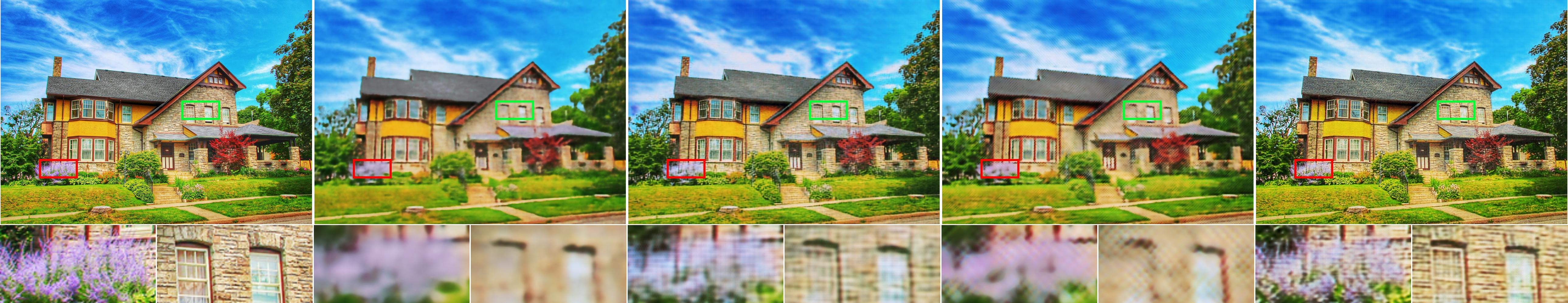}
\leftline{\qquad\  \textbf{Ground Truth}\qquad\quad\ \ \textbf{Skip, Iter 1000}\qquad\qquad  \textbf{Bi-Skip, Iter 1000}\qquad\quad \textbf{Skip, Iter 2400}\qquad\quad  \textbf{Bi-Skip, Iter 2400}}
\caption{Generated images at different iterations by using Skip and Bi-Skip.}
\label{fig.compare}
\end{figure*}
\subsection{Generative Adversarial Networks}
Generative Adversarial Networks is designed by Goodfellow\cite{goodfellow2014generative} to define a game between two competing networks called generator and discriminator. The generator receives the noise as an input and output a sample. By receiving the real and the generated sample as inputs, the discriminator is designed to distinguish the two inputs as much as possible. The goal of generator is to fool the discriminator by generating a convincing sample which can not been distinguished from the real one. In ideal conditions, the probability that a trained discriminator treats the input as its real is 0.5.

In image-to-image translation, GAN is known for its ability to generate convincing samples. However, the vanilla version suffers from lots of problems such as the model collapse, vanishing gradients, and etc, as described by Salimans\cite{salimans2016improved}.
The criterion of GAN is to minimize JS divergence between the data and the generator's distributions. JS divergence is 0 when the two inputs obey the uniform distribution, which would make gradients vanished. Though JS has strong divergence, its drawback mentioned above often makes the model fail to train. Arjovsky\cite{arjovsky2017wasserstein} comprehensively elaborates the problem and proposes the Wasserstein GAN (WGAN) which uses Earth-Mover distance as the criterion. WGAN uses Lipschitz constraint to clip the weights in range [-1, 1], which makes training more stable. Meantime, the use of weight clipping on the criteria can also lead to undesired behaviors. Gulrajani \cite{gulrajani2017improved} proposes an alternative to clip weights by penalizing the norm of gradients of the criteria with respect to its inputs. This method can perform better than the standard WGAN and enable stable training of wide variety of GAN with almost no hyperparameter tuning.
\subsection{Conditional Adversarial Networks}
Conditional Adversarial Networks is a net that both the generator and discriminator are conditioned on some extra information. 
Isola\cite{isola2017image} proposes to use cGAN in image-to-image translation, whose architecture is also known as pix2pix. Unlike the vanilla version of GAN, cGAN learns a mapping from an observed image $x$ and a random noise $z$ to the ground truth $y$ with $G:x,z\rightarrow{y}$. 
In the cGAN architecture, the discriminator's job remains unchanged, but the generator not only fools the discriminator but also minimizes the divergence between the generated sample and the ground truth. 
As described above, the valuable insight of cGAN makes image-to-image translation more diverse and stable. Thus, Kupyn\cite{Kupyn2017DeblurGAN} and Nah\cite{nah2017deep} use cGAN to deblur images in their models and output the very competitive results. Nah\cite{nah2017deep} proposes a multi-scale convolutional generator and presents the relative content loss to regularize the criteria of cGAN. Kupyn et al.\cite{Kupyn2017DeblurGAN} adopts the perceptual loss which is obtained by some feature extractors in the improved WGAN. All of above deblurring networks show that the cGAN that containing a specified generator has an unique strength to perform blur-to-sharp translation.
\subsection{Blur Dataset}
Conventional methods set up blur datasets by convolving the synthetic kernels on sharp images. In reality, the blurred image generated by this method is not identical with the real one. There are two convincing methods to generate blurred images. One is to record the sharp information to be integrated over time for blur image generation. Then the integrated signal is transformed into pixel values by the nonlinear Camera Response Function (CRF), such as GoPro dataset\cite{nah2017deep}. Another is to record 6D camera trajectories to generate blur kernels and then convolve them with sharp images by generating several common degradations, such as Lai et al.\cite{lai2016comparative}. Besides, the blurs of dataset\cite{kohler2012recording} are caused by replaying recorded 6D camera motion, assuming linear CRF.

\section{Analysis}
Before introducing the proposed method, we first explore the generator's ability to extract deep priors and then analyze the problem in the optimization.
\begin{figure}
\centering
\includegraphics[width=8.2cm]{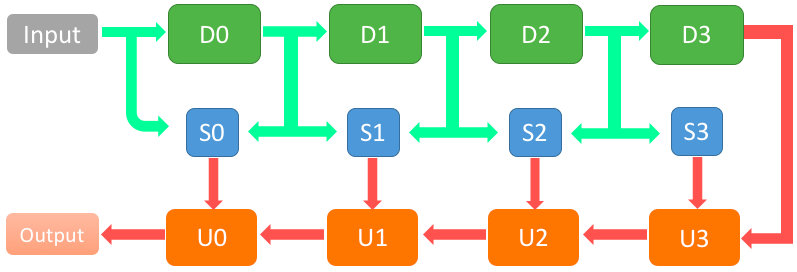}
\caption{The brief framework of Bi-Skip. This network consists of a contracting path, skipping path and expansive path. The most significant difference is that our skipping path can extract both shallow and deep priors for the same scale.} 
\label{fig.structure}
\end{figure}
\subsection{Deep Priors}
\label{sec.3-1}
Initially, U-Net is proposed to segment medical images for its excellent ability to extract useful features to characterize the contour of images. After that, Skip-Net is proposed to concatenate features of shallow and deep layers. This is a symmetric encoder-decoder that the decoder receives additional encoder features extracted by $1\times 1$ conv. This method has a prominent performance except for motion deblurring. As supposed, a generator network should own the ability to capture a great deal of image statistics priors. However, it always fail to use these networks to deblur images in some complex scenarios. In our opinion, a good generator should be capable to digest these priors rather than only to extract features.

Thus, we design a Bi-Skip network to improve the generating ability. In this section, a brief framework of the Bi-Skip is shown in figure \ref{fig.structure}. The network consists of a contracting path (D), skipping path (S) and expansive path (U). In the contracting path, the network performs down-sampling and generates shallow and deep features (shown in green arrow) for each scale. In the expansive path, a transposed convolution is utilized for up-sampling. In the skipping path, both shallow and deep features are concatenated with up-sampled features. As shown in figure \ref{fig.structure}, we present the network with 4 down-samplings and 4 up-samplings as an example.
\begin{figure}
\includegraphics[width=8.4cm]{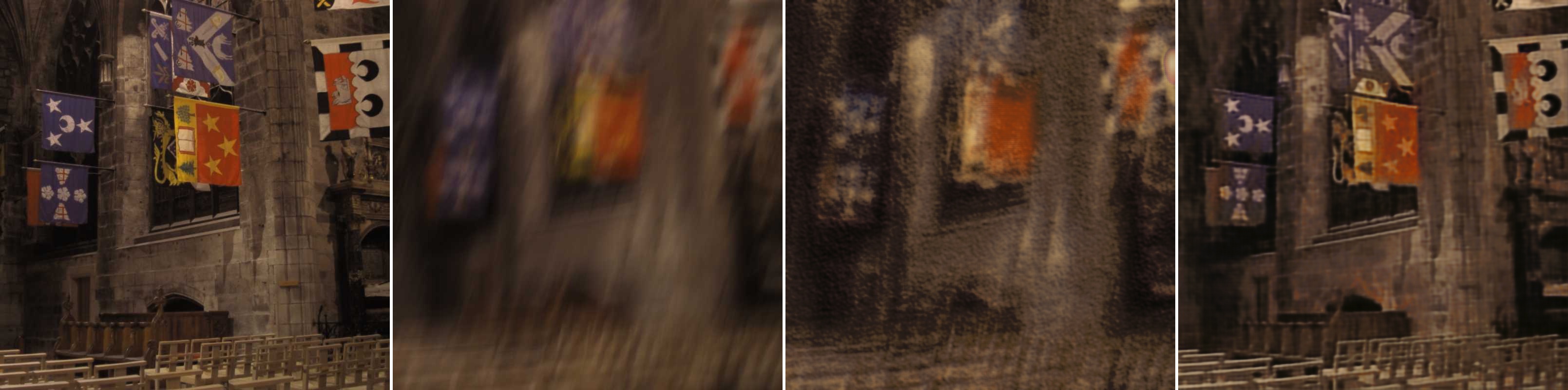}
\centering
\caption{Restore a blurred image to fit the sharp one. From left to right: sharp image, blurred image, restored images at iteration 100 and 500 by using Bi-Skip.}
\label{fig.blur_case}
\end{figure}
\begin{figure*}[t]
\centering
\includegraphics[width=17.8cm]{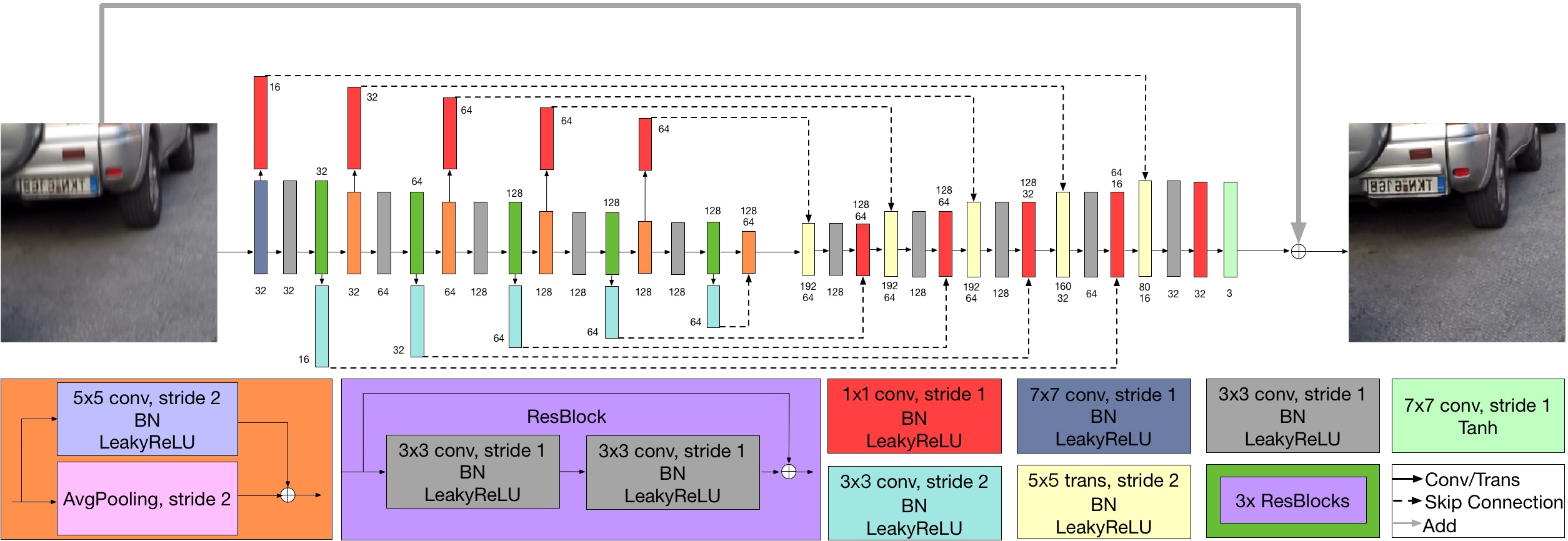}  
\caption{The architecture of proposed Bi-Skip network.}
\label{fig.framework}
\end{figure*}
To understand the characteristic of the proposed Bi-Skip network, we consider a typical reconstruction problem. Given a target image $x_0$ and a random noise $z$, the task there is to optimize parameters to reproduce the target. Without any restriction on the generated image, the objective is defined below:
\begin{equation}
   \operatorname*{min}\limits_{G} ||G(z)-x_0||^2
\end{equation}
In figure \ref{fig.compare}, we show the generation ability of Skip and Bi-Skip at iteration 1000 and 2400. Obviously, Bi-Skip can fit the target with faster speed and better learning ability. 
Also, we utilize a blurred image as input to fit the sharp one and restored images at different iterations are shown in figure \ref{fig.blur_case}.
\subsection{Problem}
\label{sec.3-2}
\begin{figure}[t]
\includegraphics[width=8.4cm]{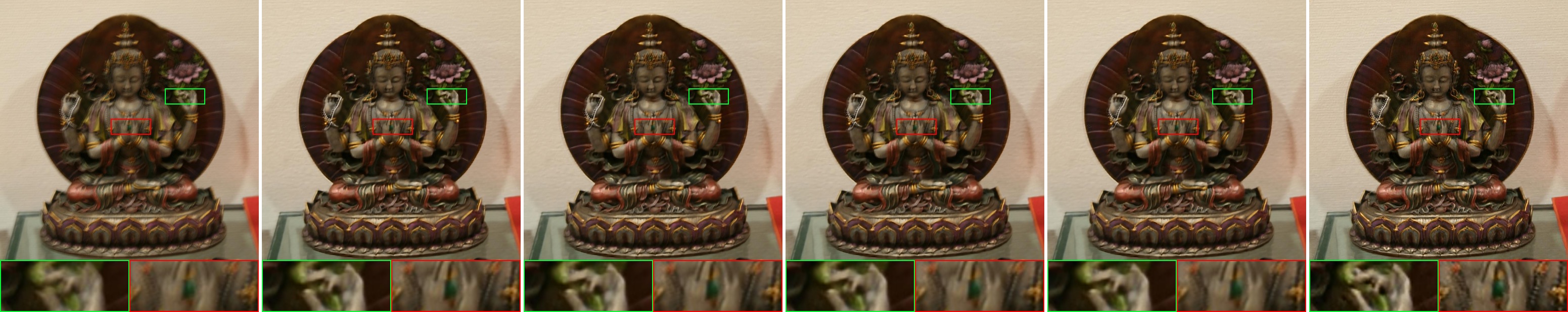}
\centering
\caption{From left to right: Blurred image, the deblurred images by Bi-Skip trained with $L_1$, $L_2$, perception, bi-level and self-paced schemes, respectively.}
\label{fig.op-problem}
\end{figure}
For image-to-image translation, the condition is added in the generator to restrict the distribution of the output. For image deblurring, Nah \cite{nah2017deep} applies MSE as the content loss and Kupyn \cite{Kupyn2017DeblurGAN} utilizes the perceptual loss \cite{johnson2016perceptual}. However, the above two conditions are not identical because the mapping about two loss spaces can not guarantee a one-to-one correspondence.
For blur-to-sharp translation, the purpose of conditions is to guide the output to obey the distribution of the sharp. The MSE constrains the output to be identical with the sharp one in both pixel and feature level. While, the perceptual loss only constrains the feature similarity. Ideally, condition should be subject to $||{G}(z)-x_0||=0$ rather than $||\mathcal{F}({G}(z))-\mathcal{F}(x_0)||=0$, where $\mathcal{F}$ is a feature extractor. To elaborate this problem, we combine the two content losses as our condition.

Besides, we find that complex motion blur often perturbs the network in the training process. Intuitive choice to solve this problem is to train the model with simple motion blur first and then to train it with complex samples. However, it is intractable to clear up the dataset according to the motion complexity. Derived from this idea, we introduce a self-paced mechanism for our networks. And figure \ref{fig.op-problem} presents the  deblurred images in different optimization schemes. '$L_1$' and '$L_2$' denote the pixel-losses regularized by $L_1$-norm and $L_2$-norm, respectively. The 'Perception' denotes the feature-loss. 
\section{Proposed Method}
\label{sec.4}
By considering the above issues, we propose a Bi-Skip network and adopt a combined condition which we call bi-level loss. At the same time, we introduce a self-paced mechanism in the training process.
\subsection{Network Architecture}
Figure \ref{fig.framework} shows the architecture of the proposed Bi-Skip. The contracting path consists of 5 down-samplings and 5 up-samplings. Specially, the down-sampling is performed by introducing the residual which is shown in orange box. Concretely, a \textsl{5$\times$5 conv} is used to obtain the residual and a \textsl{pooling} is to downsample. In each scale except for the smallest one, this path follows a \textsl{3$\times$3 conv} (gray box) and \textsl{3$\times$Resblocks} (bottle green box). The skipping path in our network is to extract both shallow and deep features. The first layer, in each scale above, follows a \textsl{1$\times$1 conv} (red box) to extract the shallow features. And the Res-Blocks follows a \textsl{3$\times$3 conv} (cyan box) to extract the deep features. Then the two features are concatenated in the expansive path. By using a \textsl{5$\times$5 transposed convolution} (yellow box), the features are up-sampled. After that, expansive path follows a \textsl{3$\times$3 conv} and a \textsl{1$\times$1 conv} in each scale. Besides, the Batch Normalization is removed from the last layer which is a residual in our network. The channels of down-sampling and up-sampling are set to [32, 64, 128, 128, 128]. Meantime, the channels are set to [16, 32, 64, 64, 64] in the skipping path for both shallow and deep features. 

And also, we propose some baselines which are brief models of Bi-SKip (\textbf{BS}). One baseline is skip network (\textbf{S}) where Res-Blocks and deep feature extractor are removed. The Res-Block only consists of 1 \textsl{conv} in \textbf{BS-cR}. The next baseline is \textbf{BS-w/o-R} that removing the Res-Blocks from \textbf{BS}. 

\subsection{Loss Function}
An adversarial loss and a content loss are combined in our method. Total loss can be formulate as follows:
\begin{equation}
   \mathcal{L}_{total} = \mathcal{L}_{adv} + \frac{1}{n}\cdot\mathcal{L}_{cont} 
\end{equation}
As we analyzed above, the feature-level loss are not identical with the pixel-level loss. We introduce a bi-level loss below:
\begin{equation}
\begin{aligned}
\mathcal{L}_{cont} = \sum_{i=1}^{n}&v_i(\frac{\gamma_1}{c_1w_1h_1}\left \| x_i - \tilde{x}_i \right \|_1 \\
  &+ \frac{\gamma_2}{c_2w_2h_2}\left \| \mathcal{F}(x_i)-\mathcal{F}(\tilde{x}_i)) \right \|_{2}^{2})
 \end{aligned}
\end{equation}
where $i$ index the current sample, $n$ is sample number, $x$ is the sharp image, $\tilde{x}$ is the generated image, $\mathcal{F}$ is a VGG19-Net which is trained on ImageNet. The purpose of $\mathcal{F}$ is to extract perceptual features in our method. The pixel-level loss is normalized by image's channel $c_1$, width $w_1$ and height $h_1$, and the feature-level loss is normalized by its channel $c_2$, width $w_2$ and height $h_2$. $\gamma_1$ and $\gamma_2$ are the weights for pixel and perception loss respectively. Additionally, $v$ is a self-paced weight to adjust the effect of current sample in the training process.

Most of the image-to-image translation works use the vanilla GAN as the loss. Further, we introduce the self-paced weight to WGN as our adversarial loss which is defined as below:
\begin{equation}
\mathcal{L}_{adv} = \operatorname*{\mathbb{E}}\limits_{x\sim \mathbb{P}_r}[D(x,v)] - \operatorname*{\mathbb{E}}\limits_{\tilde{x}\sim \mathbb{P}_g}[D(\tilde{x},v)]
\end{equation}Similar with the bi-level loss, we also use self-paced weight to adjust the discriminator how to learn. And then, the optimization mechanism is:
\begin{equation}
\operatorname*{min}\limits_{G}\operatorname*{max}\limits_{D\in\mathcal{D}}\quad\mathcal{L}_{total}
\end{equation}where $\mathcal{D}$ is the set of 1-Lipschitz function.
More recently, WGAN-GP \cite{gulrajani2017improved} is an alternative method to clip the weights in the training process of WGAN and the self-paced gradient penalty term here is:
\begin{equation}
\mathcal{P}(v,\hat{x})=\beta \operatorname*{\mathbb{E}}\limits_{\hat{x}\sim \mathbb{P}_{\hat{x}}}[(\|\nabla_{\hat{x}}D(\hat{x},v)\|-1)^2]
\end{equation}where $\hat{x} = \alpha x + (1 - \alpha)\tilde{x}$. Additionally, the task of this paper is to restore the sharp image from the blurred one and $D$ used in this work mainly focuses on the difference between two images rather than its own ability to distinguish them, so the sigmoid layer is removed from the discriminator.

\subsection{Optimization}
The weight $\mathbf{v}$ is controlled by the regime of self-paced learning in our method. However, the proposed total loss is a minimax problem and it is intractable to impose the self-paced regularizer on $\mathcal{L}_{total}$. An alternative solution is to adjust $\mathbf{v}$ by imposing the regularizer on our bi-level loss and the relative optimization is designed as below:
\begin{equation}
v_i^{*}=arg\operatorname*{min}\limits_{v_i\in[0,1]} \mathcal{L}_{cont} + f(\mathbf{v},\lambda)
\end{equation}According to the three conditions in the definition, we adopt the dynamic self-paced regularizer proposed in \cite{li2017self} and its formula is:
\begin{equation}
f(\mathbf{v},\lambda;t) = \lambda\cdot\left(\frac{1}{q(t)}\|\mathbf{v}\|_2^{q(t)}-\sum_{i=1}^{n}v_i\right)
\end{equation}Where $q(t)>1$, which is a monotonic decreasing function with respect to the iteration $t$. And we design the dynamic parameter $q(t)$ as:
\begin{equation}
\label{qt}
q(t) = tan\left[(1-\frac{1}{2}\cdot\frac{t}{T+1})\cdot\frac{\pi}{2}\right]
\end{equation}In this majorization step, the closed-form optimal solution for $v_i$ ($i=1,2,\ldots,n$) is:
\begin{equation}
v_i^*=
\left\{
\begin{aligned}
\label{v-optimize}
(1-\frac{l_i}{\lambda^t})&^{(1/(q(t)-1))}, l_i<\lambda^t\\
&0\qquad \qquad, l_i\ge\lambda^t
\end{aligned}
\right.
\end{equation}Where $l_i$ is the $i^{th}$ sample's bi-level loss and $l_i=\frac{\gamma_1}{c_1w_1h_1}\left \| x_i - \tilde{x}_i \right \|_1 + \frac{\gamma_2}{c_2w_2h_2}\left \| \mathcal{F}(x_i)-\mathcal{F}(\tilde{x}_i)) \right \|_{2}^{2}$. It is difficult to set the value of $\lambda^t$. There, we obtain the range of loss in advance and set the initial value. Concretely, we store the maximum bi-level loss among samples and set $\lambda^t$ to it, $\lambda^t=max\{l_i^{t-1}|i\in[1,2,\ldots,n]\}$. Specially, $\lambda^{t=1}$ is set to $\infty$, which is a way to select which sample to train according to the potential ability of the model. 

Once the optimal self-paced weight is given, the minimax problem of the total loss can been easily solved. The discriminator can be optimized by the following:
\begin{equation}
\label{D-optimize}
\operatorname*{min}\limits_{D\in{\mathcal{D}}} -\mathcal{L}_{adv}+\mathcal{P}(v,\hat{x})
\end{equation}And the corresponding generator can be obtained as following scheme:
\begin{equation}
\label{G-optimize}
\operatorname*{min}\limits_{G}-\operatorname*{\mathbb{E}}\limits_{G(x)\sim\mathbb{P}_g}[D(G(x),v)] + \mathcal{L}_{cont}
\end{equation}
The total optimization steps are shown in \textbf{Algorithm 1}.
\begin{algorithm}[t]
\label{algoritOptimalityhm-1}
\caption{The self-paced regime of the proposed method}
\hspace*{0.02in}{\bf Input:} 
The training dataset $\mathcal{S}$.\\
\hspace*{0.02in}{\bf Output:}
Optimized Generator $G$ and Discriminator $D$.
\begin{algorithmic}[1]
\State Initialize $G$, $D$ and set $\lambda^{t=1}\rightarrow\infty$.
\For{$t=1$ \textbf{to} $T$}
\State Calculate $q(t)$ by $Eq.(\ref{qt}$). 
\While {$not\ converged$}
\State Calculate samples' bi-level loss $l_i$.
\State Update the self-paced weight $v_i$ by $Eq.(\ref{v-optimize})$.
\While {$in\ loop$}
\State Update $D$ by $Eq.(\ref{D-optimize})$.
\EndWhile
\State Update $G$ by $Eq.(\ref{G-optimize})$.
\EndWhile
\State Update $\lambda^{t+1}=max\{l_i|i\in[0,1]\}$.
\EndFor
\State Return $G$ and $D$.
\end{algorithmic}
\end{algorithm}

\section{Training}
Except for training the model with self-paced mechanism, we also train it with other optimization schemes which are listed in the table 1 and we set them as our baselines. Simply, there lists some shorthands to denote different schemes. And \textbf{A} denotes the adversarial loss, \textbf{1} and \textbf{2} denote the pixel-loss which is regularized by the $L_1$-norm and $L_2$-norm respectively, \textbf{P} denotes the feature-loss, \textbf{S} denotes the self-paced mechanism. By combining these singe losses, we present the relative optimization schemes. And we use Xavier \cite{glorot2010understanding} to initialize the model weights. The ratio of $D$ to $G$ is set to $2:1$. Meanwhile, the ADAM \cite{kingma2014adam} is utilized as optimizer and the initial learning rate is set to $10^{-4}$. The weight $\gamma_1$, $\gamma_2$ are set to $100$, $0.1$ respectively. The total epoch is set to 300 and the learning rate linearly decays to zero after half of the epoch. In the final training, we increase the epoch to 1000 and then obtain the optimized Bi-SKip network.
\section{Experiment}
\label{tab.1}
\begin{figure*}
\includegraphics[width=17.8cm]{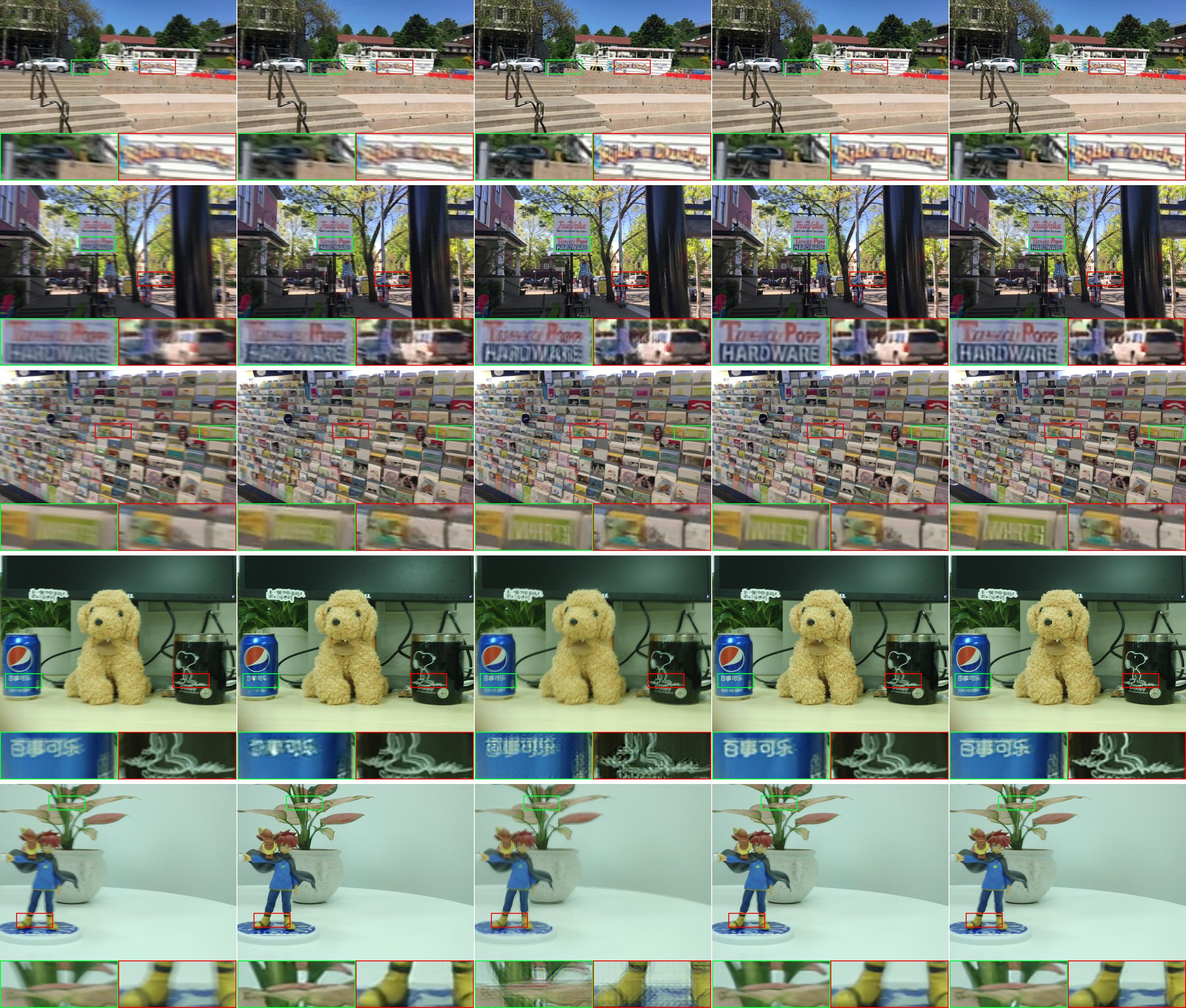}
\centering
\caption{From left to right: the blurred image, the deblurred images by Nah \cite{nah2017deep}, Kupyn \cite{Kupyn2017DeblurGAN}, Tao \cite{tao2018scale} and the proposed method, respectively.}
\label{fig.final}
\end{figure*}
\begin{figure*}
\includegraphics[width=17.8cm]{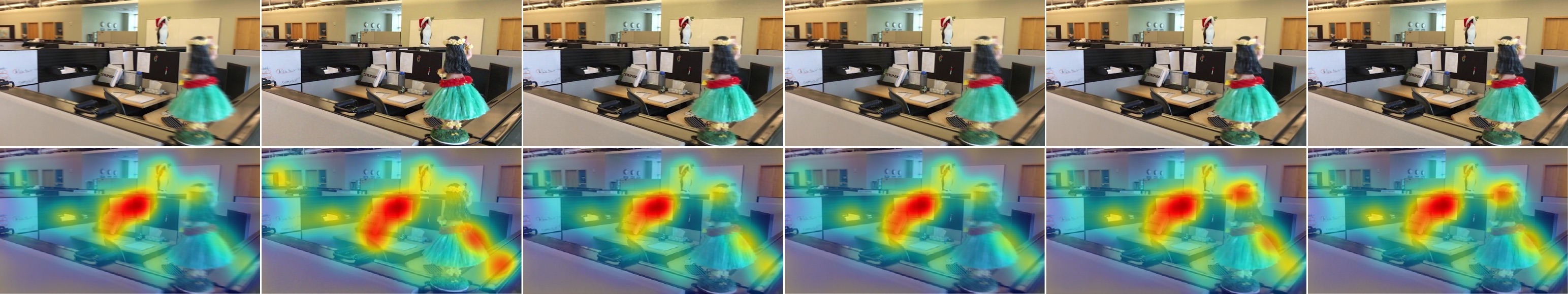}
\caption{The first row represents the blurred image, sharp image and the deblurred images obtained by Nah \cite{nah2017deep}, Kupyn \cite{Kupyn2017DeblurGAN}, Tao \cite{tao2018scale}, the proposed method, respectively. The second row represents the corresponding saliency maps.} 
\label{fig.salient-show}
\end{figure*}
\begin{table}[ht]
\tiny
\centering
\caption{The proposed method's baselines which are evaluated on the GoPro dataset.}
\begin{tabular}{c|c|c|c|c|c|c}
\hline
\hline
Model & A2-S  & A2-BS-cR & A2-BS-w/o-R & A2-BS & A2P-BS & A1P-BS         \\ \hline
PSNR  & 27.94 & 28.51    & 28.12        & 29.10  & 29.13  & 29.29          \\ \hline
SSIM  & 0.8959 & 0.9065    & 0.8992       & 0.9175 & 0.9181  & 0.9211           \\ \hline
\hline
Model & 1-BS  & 2-BS     & P-BS        & 2P-BS & AP-BS  & SA1P-BS        \\ \hline
PSNR  & 28.94 & 28.92    & 26.77       & 29.11 & 29.16  & \textbf{29.82} \\ \hline
SSIM  & 0.9145 & 0.9136    & 0.8742        & 0.9189 & 0.9196  & \textbf{0.9330} \\ \hline
\end{tabular}
\end{table}
Our experiments are performed on a single Tesla P40 GPU platform. For fair comparison, we use GoPro dataset to train our model. And the dataset contains 2,103 pairs for training and 1111 pairs for evaluation. In each iteration, we sample a batch of 1 blurred image and randomly crop 256$\times$256 patch as training input. By comparing the baselines listed in table 1, we use \textbf{A2-S}, \textbf{A2-BS-cR}, \textbf{A2-BS-w/o-R}, \textbf{A2-BS} to present the ability of \textbf{BS}. There is a obvious performance improvement when deep features are extracted in the network. And also, the rests are utilized to explore the different optimization schemes. It is effective to adopt the bi-level loss \textbf{A1P-BS} to train the model and the performance has a substantial increase when self-paced mechanism (\textbf{SA1P-BS}) is utilized.
\begin{table}[ht]
\caption{Quantitative results on GoPro and K\"ohler datasets.}
\small
\centering
\begin{tabular}{c|c|c|c|c|c}
\hline
\hline
\multirow{2}{*}{Method} & \multicolumn{2}{c|}{Gopro} & \multicolumn{2}{c|}{K\"ohler} & \multirow{2}{*}{Time} \\ \cline{2-5}
                        & PSNR        & SSIM         & PSNR         & MSSIM        &                       \\ \hline
\textit{Sun et al.}     & 24.64       & 0.8429       & 25.22        & 0.7735       & 20min                 \\ \hline
\textit{Nah et al.}     & 29.08       & 0.9135       & 26.48        & 0.8079       & 3.09s                 \\ \hline
\textit{Kupyn et al.}   & 28.7        & \textbf{0.958}        & 26.1         & 0.816        & 0.85s                 \\ \hline
\textit{Tao et al.}     & 30.26       & 0.9342       & 26.75        & 0.8370        & 1.87s                 \\ \hline
\textit{Ours}           & \textbf{30.37}       & 0.9541      & \textbf{26.98}        & \textbf{0.8407}       & \textbf{0.76s}                      \\ \hline
\end{tabular}
\end{table}
\subsection{Evaluation}
We compare our test results with methods \cite{Sun2015Learning,nah2017deep,Kupyn2017DeblurGAN,tao2018scale}on both qualitative and quantitative criteria and show the running time on a single image (table 2). Additionally, more evaluations in dataset\cite{su2017deep} and real datasets are show in figure \ref{fig.final}. K\"ohler dataset \cite{kohler2012recording} contains 4 sharp images and 48 blurred images where there exit 12 images to match each one sharp image. The blurs in this dataset are obtained by replaying recorded 6D motion trajectory. This is a benchmark for evaluating different deblurring methods and we show the quantitative results in table 2. Due to the synthetic kernels are complex in this dataset, results are not ideal as expected. Even so, the proposed method still has quite a performance.
\subsection{Saliency Based Evaluation Indicator }
Visual saliency is a process of getting a visual attention region precisely from an image. The attention is a behavioral and cognitive process of selectively concentrating on one aspect within an environment while ignoring other things. From the human eyes structure, the motion blur has a great influence on the visual attention. 

In order to highlight the superiority of our method on visual effects, we propose a pixel-aware evaluation method, which is called saliency based evaluation indicator. By using this indicator, we can understand an image content from pixel level, which accurately represents the deblurring details from human perception. We use \cite{hou2012image} to track the eye movement trajectory to highlight the visual focus. The visual comparison results are shown in figure \ref{fig.salient-show}. The first row represents the blurred image, sharp image and the deblurred images obtained by Sun\cite{Sun2015Learning}, Kupyn \cite{Kupyn2017DeblurGAN}, Tao \cite{tao2018scale}, the proposed Bi-Skip, respectively. The second row represents the corresponding saliency maps. A human can perceive the details of this scene, and will give more attentions to the Barbie doll and the stickers on the computer. And the saliency maps obtained by the eye-tracking model \cite{hou2012image} use the heat map to present the attention regions, which shows that the deeper the color, the more focus the human eyes. Two conclusions from the results of saliency maps are: (1) The fuzzy image obscures the human visual focus, which proves that the proposed debtor evaluation index is effective. (2) Compared with other methods, the results obtained by the proposed method can be capable of highlighting the visual focus and be more cater to the subjective perception of human vision on an image.
\section{Conclusion}
This paper presents an innovative approach which trains GAN to deblur the image by introducing the self-paced learning. We find that the proper generator can be used as deep priors and point out that the solution for pixel-based loss is not same with the one for perception-based loss. By using these ideas as starting points, a Bi-Skip network is proposed to improve the generating ability and a bi-level loss is adopted to solve the problem that common conditions are non-identical. Besides, considering that the complex motion blur will perturb the network in the training process, a self-paced mechanism is introduced to enhance the robustness of the network. Through extensive evaluations on both qualitative and quantitative criteria, it is demonstrated that our approach has a competitive advantage over state-of-the-art methods.
{\small
\bibliographystyle{abbrv}
\bibliography{refs}
}
\end{document}